\documentclass[12pt]{article}
\usepackage{amsmath}
\usepackage{amsthm}
\usepackage{amssymb}
\usepackage{color}
\usepackage{graphicx,psfrag,epsf}
\usepackage{enumerate}
\usepackage{natbib}
\usepackage{subfigure}
\usepackage{authblk}
\usepackage{parskip}
\usepackage{xcolor}
\usepackage{multirow}
\usepackage[linesnumbered,ruled,vlined]{algorithm2e}

\newcommand{\blind}{0}

\newcommand{\M}[1]{\boldsymbol{#1}}  
\newcommand{\V}[1]{\boldsymbol{#1}}  
\newcommand{\X}[1]{\mathbf{#1}}  
\newcommand{\Unif}[0]{\textrm{Uniform}}
\DeclareMathOperator*{\argmin}{\arg\!\min}

\newcommand{\dataset}{{\cal D}}

\DeclareMathOperator*{\minimize}{Minimize}
\DeclareMathOperator*{\supp}{supp}

\allowdisplaybreaks[1]

\SetCommentSty{mycommfont}

\SetKwInput{KwInput}{Input}                
\SetKwInput{KwOutput}{Output}              
\SetKwInput{KwInit}{Initialization}

\newtheorem{thm}{Theorem}

\begin{document}

\def\spacingset#1{\renewcommand{\baselinestretch}%
{#1}\small\normalsize} \spacingset{1}


\if0\blind
{
	{\title{\bf Variable selection for Gaussian process regression through a sparse projection
		\author{Chiwoo Park, David J. Borth, Nicholas S. Wilson, and Chad N. Hunter}
		}
	}
	\maketitle
} \fi
\if1\blind
{
	\bigskip
	\bigskip
	\bigskip
	\title{\bf Variable selection for Gaussian process regression through a sparse projection}
	\author{}
	\maketitle
} \fi

\begin{abstract}
This paper presents a new variable selection approach integrated with Gaussian process (GP) regression. We consider a sparse projection of input variables and a general stationary covariance model that depends on the Euclidean distance between the projected features. The sparse projection matrix is considered as an unknown parameter. We propose a forward stagewise approach with embedded gradient descent steps to co-optimize the parameter with other covariance parameters based on the maximization of a non-convex marginal likelihood function with a concave sparsity penalty, and some convergence properties of the algorithm are provided. The proposed model covers a broader class of stationary covariance functions than the existing automatic relevance determination approaches, and the solution approach is more computationally feasible than the existing MCMC sampling procedures for the automatic relevance parameter estimation with a sparsity prior. The approach is evaluated for a large number of simulated scenarios. The choice of tuning parameters and the accuracy of the parameter estimation are evaluated with the simulation study. In the comparison to some chosen benchmark approaches, the proposed approach has provided a better accuracy in the variable selection. It is applied to an important problem of identifying environmental factors that affect an atmospheric corrosion of metal alloys.
\end{abstract}

\noindent%
\vfill

\spacingset{1.45} 

\section{Introduction} \label{sec:intro}
Gaussian process (GP) regression is a non-parametric Bayesian approach for regression analysis \citep{rusmassen2005gaussian}. In the approach, a Gaussian process is used for defining a prior probability over an unknown regression function. The prior probability is updated with noisy observations of the function to achieve the posterior estimation of the function. It has an analytical closed form solution and has nice properties, e.g., it is the best unbiased linear predictor. The major challenges with the GP regression are its expensive computation for a large amount of data and the performance deterioration with high dimensional input variables, namely big-n and big-p issues, where $n$ stands for the number of data and $p$ stands for the input variable dimension. 

\citet{liu2017dimension} related $n$ and $p$ to the accuracy of the GP regression, based on the error bound analysis of a general kernel method \citep{fasshauer2011positive}.  According to the paper, the upper error bound of the GP regression is proportional to $n^{-1/p}$, which reduces as the number of data increases, but the reduction rate decreases as $p$ increases. That says that given the same number of data, the error bound can be larger with a larger $p$. This is the main reason for data analysts to try to reduce the input dimension by means of a dimension reduction (DR) or a variable selection (VS) technique. Another benefit of the VS is that it provides a compact subset of the input variables more relevant to the response variable of a regression analysis, so the resulting predictive model would be more interpretable. This paper is mainly concerned with the variable selection for GP regression. In Section \ref{sec:relatedworks}, we review the existing VS and DR techniques for GP to motivate our work. In Section \ref{sec:contrib}, we present our contributions and the organization of the remainder of this paper.

\subsection{Related works} \label{sec:relatedworks}
In general, a dimension reduction (DR) seeks to transform a $p$-dimensional original input $\V{x}$ to a $q$-dimensional feature $\V{z}$ for $q < p$ by a linear projection, 
\begin{equation} \label{eq:proj}
\V{z} = \M{V} \V{x}, 
\end{equation}
where $\M{V}$ is a $q \times p$ semi-orthogonal matrix with $\M{V}\M{V}^T = \M{I}$, or a non-linear projection,
\begin{equation*}
\V{z} = \M{V} \V{\phi}(\V{x}),
\end{equation*} 
for nonlinear functions, $\V{\phi}(\V{x}): \mathbb{R}^p \mapsto \mathbb{R}^{p'}$. The projection matrix $\M{V}$ is optimized with a chosen criterion. For example, a criterion of maximizing the variance of the projected features $\V{z}$ is used for the principal component analysis \citep[PCA]{hotelling1933analysis} or its nonlinear version \citep[KPCA]{scholkopf1997kernel}. As another reduction technique, the variable selection (VS) is based on a subset selection to select $q$ variables out of the $p$ variables in $\V{x}$, which can be seen as the case that the projection matrix $\M{V}$ in equation \eqref{eq:proj} is restricted to a binary matrix satisfying $\M{V}\M{V}^T = \M{I}$. The optimization of the projection matrix in DR is mostly formulated as a continuous optimization since $\M{V}$ would be a matrix of real variables. Many of the DR optimizations have analytical closed form solutions, and many others can be solved efficiently using convex optimization. Therefore, it comes with computational simplicity. However, each projected dimension in $\V{z}$ is a combination of all the variables in $\V{x}$, so the interpretation and subsequent data analysis still involves all of the original variables. In the VS, the optimization of $\M{V}$ is a combinatorial optimization problem, which is very expensive to solve. Typically, some simple greedy approaches such as the forward or backward stepwise selection are used to find a suboptimal solution, or a continuous relaxation is solved with some sparsity priors on $\M{V}$. The latter approach would give a sparse matrix $\M{V}$, so each dimension of the resulting $\V{z}$ is a function of a small subset of $\V{x}$. 

The DR or VS has been often performed as a preliminary step for a main learning task such GP regression, first performing DR or VS and then running the GP regression with reduced inputs. In earlier years, unsupervised approaches for the DR or VS were popularly applied due to simplicity. The popular unsupervised DR techniques used were principal component analysis \citep[PCA]{hotelling1933analysis}, kernel principal component analysis \citep[KPCA]{scholkopf1997kernel}, and Gaussian process latent variable method \citep[GPLVM]{lawrence2005probabilistic}. The popular VS techniques used were stepwise selection and principal variable \citep{mccabe1984principal}. A major drawback of the unsupervised approaches is that the reduced input features could be unrelated to the response variable of a regression analysis. As supervised alternatives, there are sufficient dimension reduction techniques such as the sliced inverse regression \citep[SIR]{li1991sliced}, sliced average variance estimation \citep[SAVE]{li1991sliced}, minimum average variance estimation method \citep[MAVE]{xia2009adaptive} and the gradient-based kernel dimension reduction \citep[gKDR]{fukumizu2014gradient}. For GP regression, \citet{liu2017dimension} first applied the gKDR for the dimension reduction and then performed the GP regression on the reduced dimension. Although the approaches consider the relevance to the response variable, the relevance measure is not specific to the GP regression model. 

A better approach would be to integrate the DR or VS within the GP regression, optimizing the choice of $\M{V}$ for a better GP model fit to data. One of the popular integrated approaches is based on the automated relevant determination or shortly ARD \citep{williams1996gaussian}. In the approach, the length scale parameters of a covariance function are used to determine the relevance of  input variables to the response variable. For example, a popular ARD covariance function is the squared exponential covariance function in the form of 
\begin{equation*}
\begin{split}
c_{se, ard}(\V{x}_1, \V{x}_2) = \sigma_f^2 \exp\left\{ - \sum_{j=1}^p w_j (x_{1j} - x_{2j})^2 \right\}, 
\end{split}
\end{equation*}   
where $\sigma_f^2$ is the overall variance, $x_{1j}$ is the $j$th element of the input vector $\V{x}_1$, and $w_j$ is the inverse of the length scale parameter associated with the $j$th input. The inverse length scale $w_j$ is also referred to as the relevance parameter of the $j$th input, because a smaller $w_j$ value is favored to maximize a likelihood function when the $j$th input variable is more independent of the response variable. When $w_j$ is zero, the $j$th input would have no effect on the response variable. Numerically, the likelihood maximization does not give zero $w_j$ values. For the variable selection purpose, a sparse prior can be posed to induce more zero values on the relevance parameters. Popular sparsity priors are spike and slab prior \citep{savitsky2011variable} and horseshoe prior \citep{vo2017sparse}. The resulting Bayesian variable selection requires computationally expensive Markov Chain Monte Carlo samplings. 

Another popular approach is a variable selection based on ranking input variables by its relevance to the response variable. Some KL divergence and conditional probabilities are used as a measure of the relevance. \citet{piironen2016projection} evaluated the KL divergence of the posterior distributions (of the response value) for a full GP model (containing all input variables) and a reduced model (containing a subset of the input variables). The reduced model grows iteratively through a forward stepwise selection of input variables, starting with an empty model and adding to the model one input variable every iteration that improves the KL divergence most. \citet{paananen2019variable} evaluated the relevance of each input variable to the response variable using a sensitivity measure. The sensitivity measure is defined as the degree of change in the posterior distribution of the response value under a small perturbation in each input dimension, and the degree of change is quantified by the KL divergence of the posterior distributions before and after the small perturbation. The same paper proposed another relevance measure, based on the variability of the posterior mean prediction of the response variable under a small perturbation of each input dimension. These rank measures were used to determine the relevance of the input variables to the response variable, but determining how many of the input variables are selected has not been discussed in their papers. 

There have been trials to generalize the ARD approach with a broader class of covariance forms. Please note that $c_{se, ard}$ can be written as 
\begin{equation*}
\sigma_f^2 \exp\left\{ - d_{ard}(\V{x}_1, \V{x}_2)^2 \right\}, 
\end{equation*} 
where the term, $d_{ard}(\V{x}_1, \V{x}_2) = \{(\V{x}_1 - \V{x}_2)^T \M{D} (\V{x}_1 - \V{x}_2)\}^{1/2}$, is referred to as the ARD distance, and $\M{D}$ is a diagonal matrix with $\omega_j$ be the $j$th diagonal element. The covariance $c_{se, ard}$ is the squared exponential covariance depending on the ARD distance. The squared exponential covariance can be replaced with other stationary covariance functions, creating a collection of different covariance functions that depend on the ARD distance, 
\begin{equation*}
c_{ard}(\V{x}_1, \V{x}_2) = c_{iso}(d_{ard}(\V{x}_1, \V{x}_2)),
\end{equation*}
where $c_{iso}$ is a stationary covariance function including the exponential covariance and Matérn covariances. Moreover, the ARD distance can be generalized to a more flexible form. \citet[Chapter 5]{rusmassen2005gaussian} discussed in his book the factor analysis distance, $d_{fa}(\V{x}_1, \V{x}_2) = \{(\V{x}_1 - \V{x}_2)^T (\M{\Lambda}\M{\Lambda}^T + \M{D}) (\V{x}_1 - \V{x}_2)\}^{1/2}$, where $\M{\Lambda}$ is a $p \times q$ matrix, $q < p$, and $\M{D}$ is a $p$-dimensional diagonal matrix of positives, and the distance can be combined with a stationary covariance function to define a new covariance model,
\begin{equation*}
c_{fa}(\V{x}_1, \V{x}_2) = c_{iso}(d_{fa}(\V{x}_1, \V{x}_2)).
\end{equation*}
The authors stated that the $q$ columns of $\M{\Lambda}$ could identify a few projection directions of the original inputs that are highly relevant to the response variable. However, there is an identifiability issue with $\M{\Lambda}$, because $\M{\Lambda}\M{O}$ for an arbitrary orthonormal matrix $\M{O}$ (including all rotation matrices) would achieve the same distance, and there are infinitely many versions of $\M{\Lambda}\M{O}$ with different column directions that achieve the same factor distance. \citet{tripathy2016gaussian} proposed the active subspace distance, $d_{as}(\V{x}_1, \V{x}_2) = \{(\V{x}_1 - \V{x}_2)^T \V{V}^T \M{D} \V{V} (\V{x}_1 - \V{x}_2)\}^{1/2}$, where $\V{V}$ is a $q \times p$ projection matrix with $\V{V}\M{V}^T = \M{I}$ and $\M{D}$ is a diagonal matrix of positives. In this parameterization, the projection matrix $\M{V}$ defines a low dimensional project of the input features, and the diagonal matrix defines the weights on the input features. When the diagonal elements of $\M{D}$ are all distinct, the columns of the matrix $\M{V}$ are uniquely identified. The authors combined the Matérn 32 covariance with the active subspace distance. The iterative optimization for $\M{V}$ and $\M{D}$ is proposed based on the marginal likelihood maximization criterion. Since $\M{V}$ is an orthogonal matrix, optimizing for $\M{V}$ involves a complex orthogonality-preserving iteration based on the Cayley transform \citep{wen2013feasible}. This approach is useful for the DR. Sparsifying $\M{V}$ for the VS while preserving the orthogonality is not straightforward. 

\subsection{Our contribution and the organization of the paper} \label{sec:contrib}
In this paper, we consider a stationary covariance that depends on the distance between the sparse projections of the original inputs in the form of 
\begin{equation*}
c_{S}(\V{x}_1, \V{x}_2) = c_{iso}(d_{S}(\V{x}_1, \V{x}_2))
\end{equation*}
with $d_{S}(\V{x}_1, \V{x}_2)$ being the L2 distance between the projections of the two inputs $\V{x}_1$ and $\V{x}_2$, 
\begin{equation*}
d_{S}(\V{x}_1, \V{x}_2) = ||\M{S}\V{x}_1 - \M{S}\V{x}_2||_F \mbox{ or equivalently } \{(\V{x}_1 - \V{x}_2)^T \M{S}^T \M{S} (\V{x}_1 - \V{x}_2)\}^{1/2},
\end{equation*}
where $\M{S}$ is a projection matrix, and $||\cdot||_F$ is the Frobenius norm. Unlike in the active subspace covariance $c_{as}$. The projection matrix $\M{S}$ is not required to be right-orthogonal, i.e., $\M{S}\M{S}^T$ is not necessarily an identity matrix, and it is not required to be an upper trapezoidal Cholesky factor. Without the orthogonality or upper triangularity constraint, the projection matrix is unidentifiable like in $c_{fa}$, because $\M{O}_q\M{S}$ for an arbitrary orthonormal matrix $\M{O}_q$ gives the same distance. We search for the most sparse projection matrix among infinitely many versions of $\M{O}_q\M{S}$, which would gives a sparse projection of the original inputs, so the projected features $\M{S}\V{x}$ would be a linear combination of very few original input variables. Since $\M{S}$ does not involve complex constraints such as orthogonality, optimizing the matrix would be simpler. We propose a numerical optimization for jointly optimizing the sparse $\M{S}$ and other covariance parameters. The new numerical algorithm is based on a forward stagewise approach with embedded gradient descent steps to complement the limited convergence of the coordinate descent steps of the forward stagewise for non-convex objective functions. 

The remainder of the paper is organized as follows. Section \ref{sec:method} entails a new modeling approach for a sparse projection of the input variables in GP regression and the numerical optimization to estimate the model parameters. Section \ref{sec:sim} analyzes the numerical performance of the new approach with a comprehensive set of simulated scenarios, comparing it to the results from some chosen benchmark approaches. Section \ref{sec:data} shows the numerical performance of the new approach with a motivating example of identifying environmental factors affecting atmospheric corrosion of a metal alloy. We conclude this paper in Section \ref{sec:conc}.

\section{GP regression with a sparse low-rank projection} \label{sec:method}
Consider a general regression problem of estimating an unknown regression function $f$ that relates a $p$-dimensional input $\V{x} \in \mathbb{R}^p$ to a real response $y$, using noisy observations $\dataset = \{(\V{x}_i,y_i), i=1,\ldots, N\}$,
\begin{equation*}
y_i = f(\V{x}_i) + \epsilon_i, \qquad i =1, \dots, N,
\end{equation*}
where $\epsilon_i \sim \mathcal{N}(0, \sigma^2)$ is white noise, independent of $f(\V{x}_i)$. In the GP regression, the underlying regression function $f$ is assumed a realization of Gaussian process with zero mean and covariance function $c_S$. Here we limit the covariance function to be stationary, which implies that the covariance between two function values, $f(\V{x}_i)$ and $f(\V{x}_j)$, depends on the distance $d_S$ between $\V{x}_i$ and $\V{x}_j$, 
\begin{equation*}
c_S(\V{x}_i, \V{x}_j) = c_{iso}(d_S(\V{x}_i, \V{x}_j); \V{\theta}),
\end{equation*}
where $c_{iso}$ is a stationary covariance, and $\V{\theta}$ is the parameter(s) of the stationary covariance. The distance $d_S$ over $\mathbb{R}^p$ is defined in the following quadratic form, 
\begin{equation*}
d_S(\V{x}_i, \V{x}_j) = \{(\V{x}_i-\V{x}_j)^T \M{Q} (\V{x}_i-\V{x}_j) \}^{1/2},
\end{equation*}
where the $p \times p$ matrix $\M{Q}$ should be positive semidefinite for $d_S$ being a proper distance satisfying positivity and triangle inequality. The distance $d_S$ is referred to as the Mahalanobis distance or generalized L2 distance in the literature \citep{chandra1936generalised}. Please note that the existing generalized ARD approaches used the same form of a covariance function with different parameterizations of $\M{Q}$, e.g., the low-rank factorization $\M{Q} = (\M{\Lambda}\M{\Lambda}^T + \M{D})$ with a $p \times q$ matrix $\M{\Lambda}$ and a diagonal matrix $\M{D}$ in \citet{rusmassen2005gaussian} and the spectral decomposition $\M{Q} = \V{V}^T \M{D} \V{V}$ with a $q \times p$ right-orthogonal matrix $\M{V}$ in \citet{tripathy2016gaussian}. As we discussed in the introduction, the low-rank factorization has an identifiability issue, and the spectral decomposition incurs a complexitiy in optimizing $\M{V}$ while preserving the orthogonality. In this paper, we consider a simpler parameterization, 
\begin{equation} \label{eq:facQ}
\M{Q} = \M{S}^T \M{S},
\end{equation}
where $\M{S}$ is a $q \times p$ real matrix, not required to be an orthogonal matrix or an upper trapezoidal triangular Cholesky factor. Without the orthogonality and upper triangularity constraints, the matrix $\M{S}$ is not uniquely identified as in the low-rank factorization used by \citet{rusmassen2005gaussian}, because $\M{O}_q \M{S}$ for an arbitrary $q \times q$ orthonormal matrix $\M{O}_q$ also gives the same form of the factorization. Among infinitely many $\M{S}$ that factorizes $\M{Q}$ in the form of \eqref{eq:facQ}, we seek a sparse factor $\M{S}$ that satisfies
\begin{equation}
\mathcal{R}(\M{S}) \le \mu,
\end{equation} 
where $\mathcal{R}$ is the sparsity norm on $\M{S}$, i.e., the $r$-norm for $r \le 1$. We have two reasons for placing the sparsity constraint. The sparsity constraint resolves the identifiability issue, and more importantly it is hoped that $d_S$ only depends on a very few variables of the $p$ original inputs for a better interpretation of the GP regression result. Please note that with the factorization, the distance $d_S$ can be written as
\begin{equation*}
d_S(\V{x}_i, \V{x}_j) = || \M{S} (\V{x}_i - \V{x}_j) ||_F,
\end{equation*}
where $||\cdot||_F$ is the Frobenius norm. The matrix $\M{S}$ projects the $p$ original inputs to $q$ dimensional features. If the projection matrix is sparse, one can have each of the projection features a linear combination of only very few original inputs. Below we propose a numerical optimization for jointly optimizing the sparse $\M{S}$ and other covariance parameters.

For describing the solution approach to optimize the parameters, we introduce a common set of notations. We denote the collection of observed input locations, $\X{X} = [\V{x}_1,\dots,\V{x}_N]^T$, and we denote the collection of observed response variables, $\V{y} =  [y_1, \ldots, y_N]^T$. With the Gaussian process prior, the prior distribution of $\V{f} = [f(\V{x}_1), \ldots, f(\V{x}_N)]^T$ is the multivariate normal distribution,
\begin{equation*}
\V{f}|\V{X}, \M{S}, \V{\theta} \sim
\mathcal{N}\left( \V{0}, \M{C}_{\M{S}, \V{\theta}} \right),
\end{equation*}
where $\M{C}_{\M{S}, \V{\theta}}$ is an $N \times N$ matrix with $(i,j)^{th}$ entry $c_S(\V{x}_i, \V{x}_j)$. The conditional distribution of $\V{y}$ is 
\begin{equation*}
\V{y}|\V{f}, \sigma^2 \sim
\mathcal{N}\left( \V{f}, \sigma^2\M{I} \right).
\end{equation*}
Let $\V{\phi}_C = \{\V{\theta}, \sigma^2\}$ to represent a set of the covariance parameters and the noise variance parameter. The marginal distribution of $\V{y}$ given $\M{X}, \V{\phi}_C$ and the distance parameter $\M{S}$ can be derived as a multivariate normal distribution,
\begin{equation*}
\V{y}|\V{X}, \M{S}, \V{\phi}_C \sim
\mathcal{N}\left( \V{0}, \sigma^2\M{I} + \M{C}_{\M{S}, \V{\theta}} \right).
\end{equation*}
The parameter set, $\V{\phi}_C$ and $\M{S}$, are jointly optimized by minimizing the negative log likelihood function,
\begin{equation} 
\mathcal{L}(\M{S}, \V{\phi}_C) = \frac{1}{2} \V{y}^T (\sigma^2\M{I} + \M{C}_{\M{S}, \V{\theta}})^{-1} \V{y} + \frac{1}{2} \log |\sigma^2\M{I} + \M{C}_{\M{S}, \V{\theta}}|
\end{equation} 
under a sparsity constraint on $\M{S}$, 
\begin{equation*}
\mathcal{R}(\M{S}) \le \mu
\end{equation*}
or equivalently its Lagrange relaxation is solved 
\begin{equation} \label{opt:RML}
\begin{split}
\minimize & \quad \mathcal{L}(\M{S}, \V{\phi}_C) + \lambda \mathcal{R}(\M{S}),
\end{split}
\end{equation} 
where $\lambda > 0$ is the Lagrange multiplier. The solution depends on a choice of two tuning parameters, the rank parameter $q$ and the sparsity parameter $\lambda$. We will discuss a numerical optimization of problem \eqref{opt:RML} for a choice of the tuning parameters in Section \ref{sec:opt}, and some technical details of the optimization are in Section \ref{sec:detail}. The choice of the tuning parameters will be covered in Section \ref{sec:hyper}. 

\subsection{\texttt{FSEG}: Forward stagewise with embedded gradient descent steps for parameter estimation} \label{sec:opt}
In this section, we present a numerical approach to solve problem \eqref{opt:RML} for estimating the covariance parameter $\V{\phi}_C$ and the distance $\M{S}$ jointly. The objective function of the problem consists of the likelihood term $\mathcal{L}$ and the $r$-norm sparsity penalty term $\mathcal{R}$. A sparsity penalized optimization problem has been studied in different problem settings. When the likelihood term is in a quadratic form and the penalty term is a $1$-norm, the problem is known as the \textit{Lasso} problem. The forward stagewise regression algorithm was quite successful for solving the Lasso problem \citep{efron2004least}. The approach is later generalized by \citet{zhao2007stagewise} for a convex likelihood term (or empirical loss) with the 1-norm penalty (BLasso) and a convex likelihood with a convex penalty function (the generalized BLasso). The major advantage of using the forward stagewise regression is that it generates the solution path containing the solutions over a wide range of $\lambda$ values, so the selection of the sparsity parameter $\lambda$ can be done by evaluating the solutions in the path with a model selection criterion. All of the convergence proofs in the existing works are based on the strong convexity assumption on the objective function including the likelihood and penalty term. For our problem \eqref{opt:RML}, the likelihood term is non-convex, so the convergence results in the past works are not applicable. In general, the forward stagewise and its variants belong to a steepest coordinate descent method, which does not provide a guarantee to converge to a local optimality for general noncovex objection functions \citep{nutini2015coordinate}, for which a gradient descent method with the full gradient provides a better convergence. However, the gradient descent numerically does not give a sparse solution even with a large $\lambda$ value, and a numerical truncation of the outcome is necessary. Here we propose a combination of the forward stagewise approach with a gradient descent method, which basically runs the forward stagewise iterations with embedded gradient steps to complement the limited convergence of the coordinate decent steps. The approach inherits the good features of the conventional forward stagewise approaches, i.e., providing the solution path for different $\lambda$ values. The new approach is referred to as the \textit{forward stagewise with embedded gradient descent step} or shortly \texttt{FSEG}.

To describe the approach, let $\V{\phi}$ denote a large vector concatenating the elements of $\M{S}$ and $\V{\phi}_C$ with its initial $q \times p$ elements from $\M{S}$ and the remaining elements from $\V{\phi}_C$, and let $J$ denote the total number of the elements in the large vector. Consider a problem of finding $\V{\phi}$ that minimizes 
\begin{equation} \label{eq:prob}
\begin{split}
\minimize_{\V{\phi} \in \mathbb{R}^{J}} \quad \Gamma(\V{\phi}; \lambda) = \mathcal{L}(\M{S}, \V{\phi}_C) + \lambda \mathcal{R}(\M{S}),
\end{split}
\end{equation}
where $\mathcal{L}$ is a non-convex function, and $\mathcal{R}$ is concave. We like to generate the solution path of the problem, including the local minimum of $\Gamma(\V{\phi}; \lambda)$ for each value of $\lambda$ ranging from 0 to infinity, where the solution path implies a series of the solutions of problem \eqref{eq:prob},  
\begin{equation*}
(\V{\phi}^{(t)}, \lambda^{(t)}; t = 1,2,\ldots),
\end{equation*}
where $\V{\phi}^{(t)}$ denotes the $t$th solution achieved with $\lambda=\lambda^{(t)}$. The initial solution $\V{\phi}^{(0)}$ is set to one obvious minimum, $\M{S} = \M{0}$ and $\V{\phi}_C = \argmin_{\V{\phi}_C'} \mathcal{L}(\M{0}, \V{\phi}_C')$ for $\lambda^{(0)} = \infty$. We start with the initial solution, and update the solution iteratively to other solutions, using the following forward stagewise steps. A forward stagewise regression belongs to a coordinate descent algorithm, which iteratively updates the solution along a chosen coordinate direction with a small step size $\epsilon$. A coordinate descent step can be written as
\begin{equation*}
\V{\phi}^{(t+1)} = \V{\phi}^{(t)} + s \V{e}_{j},
\end{equation*}
where $|s| = \epsilon$, and $\V{1}_{j}$ is a $J \times 1$ vector of all zeros except for the $j$th element being one. The $j$ indicates the variable to be updated, and $s$ defines the direction and magnitude of the update. First try the coordinate descent direction on $\Gamma$ for $(j, s)$, 
\begin{equation*}
(j_b, s_b) = \argmin_{j \in \{1,\ldots, J\}, |s|=\epsilon} \Gamma(\V{\phi}^{(t)} + s \V{e}_{j}; \lambda^{(t)}).
\end{equation*} 
Following this coordinate descent direction would make an improvement of $\Gamma$ by making a little change in one coordinate of $\M{S}$ or $\V{\phi}_C$. If the improvement is more than or equal to a small tolerance parameter $\xi$, 
\begin{equation} \label{eq:conv}
\Gamma(\V{\phi}^{(t)}; \lambda^{(t)}) - \Gamma(\V{\phi}^{(t)} + s_b \V{e}_{j_b}; \lambda^{(t)}) \ge \xi, 
\end{equation}
we take the coordinate direction to update the solution,
\begin{equation} \label{eq:back}
\V{\phi}^{(t+1)} = \V{\phi}^{(t)} + s_b \V{e}_{j_b},
\end{equation}
and keep $\lambda^{(t+1)} = \lambda^{(t)}$. Otherwise, $\Gamma$ can only be very little reduced along any coordinate directions for the current $\lambda$ value. This implies one of two scenarios, (1) the iteration is close to a local minimum of $\Gamma$ for the current $\lambda$, or (2) $\Gamma$ would not improve along any of the coordinate directions, although the current solution $\V{\phi}^{(t)}$ is far from a local minimum, i.e., the coordinate descent steps were stuck in the middle of the path to a local minimum. The latter case may happen for non-convex objective functions because the direction of the update in a coordinate descent step is restricted to one coordinate direction at a time, and any of the coordinate directions may not give any improvement in $\Gamma$, for which the coordinate decent steps simply stops possibly before reaching to a local minimum. To escape from being stuck, we relax the improvement direction from the coordinate-wise direction to the support-limited gradient by running one gradient descent step,
\begin{equation} \label{eq:grad}
\V{\phi}^{(t+1/2)} = \V{\phi}^{(t)} - \epsilon_g \nabla_{\supp} \Gamma(\V{\phi}^{(t)}; \lambda^{(t)}),
\end{equation}
where $\nabla_{\supp} \Gamma(\V{\phi}^{(t)})$ is the support-limited gradient of $\Gamma$ evaluated at $\V{\phi}^{(t)}$, and the step size $\epsilon_g$ can be chosen using a line search. Here `support' implies the support of the solution, $\supp(\V{\phi}^{(t)}) = \{j = \{1, \ldots, J\}; \V{e}_j^T \V{\phi}^{(t)} \neq 0\}$, and the `support-limited' implies that the $j$th element of the gradient vector $\nabla_{\supp} \Gamma(\V{\phi}^{(t)}; \lambda^{(t)})$ is shrink to zero if $j \notin \supp(\V{\phi}^{(t)})$; more details of the support-limited gradient can be found in Section \ref{sec:detail}. This support-limited update finds the update along a combination of the multiple coordinates belonging to the support, instead of one coordinate direction, so it finds improvement directions that are not considered in the coordinate descent. On the other hand, the support of the solution with the update remains same as that of $\V{\phi}^{(t)}$, so the sparsity is maintained unlike in the conventional gradient descent with the full gradient. If the result of the gradient step satisfies
\begin{equation} \label{eq:conv2}
\Gamma(\V{\phi}^{(t)}; \lambda^{(t)}) - \Gamma(\V{\phi}^{(t+1/2)}; \lambda^{(t)}) \ge \xi, 
\end{equation}
we take the result, 
\begin{equation} \label{eq:back2}
\V{\phi}^{(t+1)} = \V{\phi}^{(t+1/2)}.
\end{equation}
Otherwise, the $\Gamma$ value cannot be further reduced with the current $\lambda$ value. We take one forward step for reducing $\lambda$ unless the $\lambda$ value cannot be further reduced, i.e., $\lambda^{(t)}=0$, for which we stop the iteration. Choose the coordinate descent direction on the non-penalized likelihood term $\mathcal{L}$ among the first $q \times p$ coordinates of $\V{\phi}$,
\begin{equation*}
(j_f, s_f) = \argmin_{j \in \{1,\ldots, q \times p\}, |s|=\epsilon} \Gamma(\V{\phi}^{(t)} + s \V{e}_{j}; \lambda=0),
\end{equation*} 
and take the forward stagewise step,
\begin{equation} \label{eq:for}
\V{\phi}^{(t+1)} = \V{\phi}^{(t)} + s_f \V{e}_{j_f}.
\end{equation}

\begin{algorithm}[ht!]
	\DontPrintSemicolon
	
	\KwInput{rank $q$, step size $\epsilon$, tolerance parameter $\xi$, maximum number of iterations $T_{max}$ }
	\KwOutput{solution path  $\{ (\M{\phi}^{(t)}, \lambda^{(t)}): t = 1,\ldots, T_{max}\}$}
	{\textbf{Initialization.} Set the initial solution $\V{\phi}^{(0)}$ with $\M{S}^{(0)} = \M{0}$ and
		$$\V{\phi}_C^{(0)} = \argmin_{\V{\phi}_C \in \mathbb{R}^{J-qp}} \mathcal{L}(\M{S}^{(0)}, \V{\phi}_C).$$
		Set the initial value of $\lambda$ to $\lambda^{(0)} = \infty$.}\\
	\For{t = 1:$T_{max}$}
	{
		Find the coordinate descent direction on $\Gamma$,
		$$(j_b, s_b) = \argmin_{j \in \{1,\ldots, J\}, |s|=\epsilon} \Gamma(\V{\phi}^{(t)} + s \V{e}_{j}; \lambda^{(t)}).$$
		
		\If{$\Gamma(\V{\phi}^{(t)}; \lambda^{(t)}) - \Gamma(\V{\phi}^{(t)} + s_b \V{e}_{j_b}; \lambda^{(t)}) \ge \xi$}
		{
			Update $\V{\phi}^{(t+1)} = \V{\phi}^{(t)} + s_b \V{e}_{j_b}$.\\
			Update $\lambda^{(t+1)} = \lambda^{(t)}$.
		}
		\Else
		{
			\If{$\lambda^{(t)} = 0$}
			{
				Stop the procedure.
			}
			\Else
			{
				Take the support-limited gradient descent step \eqref{eq:grad} to achieve $\V{\phi}^{(t+1/2)}$.\\
				If $\Gamma(\V{\phi}^{(t)}; \lambda^{(t)}) - \Gamma(\V{\phi}^{(t+1/2)}; \lambda^{(t)}) \ge \xi$, set $\V{\phi}^{(t+1)} = \V{\phi}^{(t+1/2)}$ and continue.
				Otherwise, find the coordinate descent direction on $\mathcal{L}$,
				$$(j_f, s_f) = \argmin_{j \in \{1,\ldots, q \times p\}, |s|=\epsilon} \Gamma(\V{\phi}^{(t)} + s \V{e}_{j}; \lambda=0),$$
				\quad update $\V{\phi}^{(t+1)} = \V{\phi}^{(t)} + s_b \V{e}_{j_b},$  \\
				\quad and update $\lambda^{(t+1)} = \min\left\{\lambda^{(t)},  \frac{\mathcal{L}(\M{S}^{(t)}, \V{\phi}_C^{(t)}) - \mathcal{L}(\M{S}^{(t+1)}, \V{\phi}_C^{(t+1)}) - \xi}{ \mathcal{R}(\M{S}^{(t+1)}) - \mathcal{R}(\M{S}^{(t)})} \right\}$.
			}
		}
	}
	\caption{forward stagewise with embedded gradient descent step (FSEG)} \label{alg:forwardstage}
\end{algorithm}

The $\lambda$ value changes from $\lambda^{(t)}$ to 
\begin{equation} \label{eq:lam_upd}
\lambda^{(t+1)} = \min\left\{\lambda^{(t)},  \frac{\mathcal{L}(\M{S}^{(t)}, \V{\phi}_C^{(t)}) - \mathcal{L}(\M{S}^{(t+1)}, \V{\phi}_C^{(t+1)}) - \xi}{ \mathcal{R}(\M{S}^{(t+1)}) - \mathcal{R}(\M{S}^{(t)})} \right\},
\end{equation}
where $\M{S}^{(t)}$ and $\V{\phi}_C^{(t)}$ are the corresponding parts of $\V{\phi}^{(t)}$. The whole algorithm is summarized in \textbf{Algorithm} \ref{alg:forwardstage}. 

The algorithm draws a monotone sequence of the solutions in terms of the values $\Gamma(\V{\phi}; \lambda)$ with variable step greater than or equal to $\xi$ as described in Theorem \ref{thm:mono}. 
\begin{thm} \label{thm:mono}
	For any iteration $t$, the objective function $\Gamma$ value is improved by at least $\xi$, $$\Gamma(\V{\phi}^{(t+1)}; \lambda^{(t+1)}) \le \Gamma(\V{\phi}^{(t)}; \lambda^{(t)}) - \xi.$$
	\begin{proof}
		It is obvious to show either of the updates \eqref{eq:back} and \eqref{eq:back2} with $\lambda^{t+1} = \lambda^{(t)}$ satisfies $\Gamma(\V{\phi}^{(t+1)}; \lambda^{(t+1)}) \le \Gamma(\V{\phi}^{(t)}; \lambda^{(t)}) - \xi.$ Here we show that the condition holds for the update \eqref{eq:for} with the lambda update \eqref{eq:lam_upd}. If $\frac{\mathcal{L}(\M{S}^{(t)}, \V{\phi}_C^{(t)}) - \mathcal{L}(\M{S}^{(t+1)}, \V{\phi}_C^{(t+1)})}{ \mathcal{R}(\M{S}^{(t+1)}) - \mathcal{R}(\M{S}^{(t)})} \le  \lambda^{(t)}$ in the lambda update, 
		$$\lambda^{(t+1)} = \frac{\mathcal{L}(\M{S}^{(t)}, \V{\phi}_C^{(t)}) - \mathcal{L}(\M{S}^{(t+1)}, \V{\phi}_C^{(t+1)}) - \xi}{ \mathcal{R}(\M{S}^{(t+1)}) - \mathcal{R}(\M{S}^{(t)})} \le  \lambda^{(t)},$$ and
		\begin{equation*}
		\Gamma(\V{S}^{(t+1)}; \lambda^{(t+1)}) =  \Gamma(\V{S}^{(t)}; \lambda^{(t+1)}) - \xi \le \Gamma(\V{S}^{(t)}; \lambda^{(t)}) - \xi.
		\end{equation*}
		Otherwise, i.e. if $\frac{\mathcal{L}(\M{S}^{(t)}, \V{\phi}_C^{(t)}) - \mathcal{L}(\M{S}^{(t+1)}, \V{\phi}_C^{(t+1)})-\xi}{ \mathcal{R}(\M{S}^{(t+1)}) - \mathcal{R}(\M{S}^{(t)})} >  \lambda^{(t)}$, 
		\begin{equation*}
		\Gamma(\V{S}^{(t+1)}; \lambda^{(t)}) < \Gamma(\V{S}^{(t)}; \lambda^{(t)}) - \xi.
		\end{equation*}
		Since $\lambda^{(t+1)} = \lambda^{(t)}$, 
		\begin{equation*}
		\Gamma(\V{S}^{(t+1)}; \lambda^{(t+1)}) < \Gamma(\V{S}^{(t)}; \lambda^{(t+1)}) - \xi.
		\end{equation*}
	\end{proof}
\end{thm}
Since the solution sequence is monotone with respect to the corresponding objective value and the objective function is bounded below by zero, the sequence converges in a finite number of iterations by the bounded convergence theorem. 

\subsection{Technical Details: Embedded Gradient Descent Step} \label{sec:detail}
In this section, we describe more details of the support limited gradient descent step in Line 11 of Algorithm \ref{alg:forwardstage}. The gradient descent step follows a general gradient descent step in the form of
\begin{equation*}
\V{\phi}^{(t+1/2)} = \V{\phi}^{(t)} - \epsilon_g \nabla_{\supp} \Gamma(\V{\phi}^{(t)}; \lambda^{(t)}).
\end{equation*}
A peculiarity is that the gradient used is the support-limited gradient, $\nabla_{\supp} \Gamma(\V{\phi}^{(t)}; \lambda^{(t)})$. To formally describe the gradient, we denote the $j$th coordinate of $\V{\phi}$ by $\phi_j$ and the $j$th coordinate of $\V{\phi}^{(t)}$ by $\phi_j^{(t)}$. The support-limited gradient is a $J \times 1$ vector, and its $j$th element is the first order partial derivative of $\Gamma$ with respect to $\phi_j$ evaluated at $\V{\phi} = \V{\phi}^{(t)}$ and $\lambda = \lambda^{(t)}$ if $\phi_j^{(t)} \neq 0$,
\begin{equation*}
\frac{\partial \Gamma}{\partial \phi_j} \left|_{\V{\phi} = \V{\phi}^{(t)}, \lambda = \lambda^{(t)}}\right.,
\end{equation*} 
and its $j$th element is zero if $\phi_j^{(t)} = 0$. The first order partial derivative of $\Gamma$ with respect to each $\phi_j$, 
\begin{equation*}
\begin{split}
\frac{\partial \Gamma}{\partial \phi_j}  & =  - \frac{1}{2}\V{y}^T \left( \sigma^2\M{I} + \M{C}_{\M{S}, \V{\theta}} \right)^{-1} \frac{\partial  \left( \sigma^2\M{I} +  \M{C}_{\M{S}, \V{\theta}} \right) }{\partial \phi_j}  \left( \sigma^2\M{I} + \M{C}_{\M{S}, \V{\theta}} \right)^{-1} \V{y} \\
& \quad + \frac{1}{2} tr\left[ \left(\sigma^2\M{I} +  \M{C}_{\M{S}, \V{\theta}} \right)^{-1} \frac{\partial  \left( \sigma^2\M{I} + \M{C}_{\M{S}, \V{\theta}} \right)}{\partial \phi_j}   \right] \\
& \quad + \lambda \frac{\partial \mathcal{R}}{\partial \phi_j} 
\end{split}
\end{equation*}
If $\phi_j = S_{lm}$,  
\begin{equation*}
\begin{split}
\frac{\partial  \left( \sigma^2\M{I} + \M{C}_{\M{S}, \V{\theta}} \right)_{ij}}{\partial S_{lm}} & = \left. \frac{\partial c_{iso}(d; \V{\theta}) }{\partial d} \right|_{d = d_S(\V{x}_i, \V{x}_j)} \frac{\partial d_S(\V{x}_i, \V{x}_j)}{\partial S_{lm}} \\
& = \left. \frac{\partial c_{iso}(d; \V{\theta}) }{\partial d} \right|_{d = d_S(\V{x}_i, \V{x}_j)} \frac{1}{2d_S(\V{x}_i, \V{x}_j)} \frac{\partial (\V{x}_i, \V{x}_j)^T \M{S}^T \M{S} (\V{x}_i, \V{x}_j)}{\partial S_{lm}} \\
& = \left. \frac{\partial c_{iso}(d; \V{\theta}) }{\partial d} \right|_{d = d_S(\V{x}_i, \V{x}_j)} \frac{1}{2d_S(\V{x}_i, \V{x}_j)} 2 (\M{S}(\V{x}_i - \V{x}_j) (\V{x}_i - \V{x}_j)^T)_{lm} \\
& = \left. \frac{\partial c_{iso}(d; \V{\theta}) }{\partial d} \right|_{d = d_S(\V{x}_i, \V{x}_j)} \frac{\V{s}_l^T (\V{x}_i - \V{x}_j) (x_{im} - x_{jm})}{d_S(\V{x}_i, \V{x}_j)},
\end{split}
\end{equation*}
where $\V{s}_l^T$ is the $l$th row vector of $\M{S}$, and $x_{im}$ is the $m$th element of $\V{x}_i$. The partial derivatives with respect to other coordinates are all dependent on the choice of $c_{iso}$. 

\subsection{Tuning Parameter Selection} \label{sec:hyper}
There are two tuning parameters, the sparsity parameter $\lambda$ and the rank parameter $q$. We first tried the Bayesian information criterion (BIC) to choose both of the parameters. For a choice of $q \in \{1, \ldots, Q_{max}\}$, the proposed FSEG would generate a solution path for a wide range of $\lambda$ values. Let $\{(\V{\phi}^{(t)}_q, \lambda^{(t)}_q); t = 1,\ldots, T_{max}\}$ denote the solution path for a choice of $q$. We evaluate the BIC for each solution in the solution path,
\begin{equation} \label{eq:BIC}
BIC(\V{\phi}^{(t)}_q, \lambda^{(t)}_q) = 2\mathcal{L}(\V{\phi}^{(t)}_q; \lambda^{(t)}_q) + ||\V{\phi}^{(t)}_q||_0\log(N),
\end{equation}
where $||\cdot||_0$ is the 0-norm. The $\lambda$ value conditioned on the given $q$ value can be chosen as $\lambda_q^{(t_q)}$, 
\begin{equation*}
t_q = \argmin_{t = 1,\ldots, T_{max}} BIC(\V{\phi}^{(t)}_q, \lambda^{(t)}_q).
\end{equation*} 
The value of the rank parameter $q$ can be chosen to 
\begin{equation*}
q^* = \argmin_{q = 1, \ldots, Q_{max}} BIC(\V{\phi}^{(t_q)}_q, \lambda^{(t_q)}_q),
\end{equation*}
and the final choice of $\lambda$ would be $\lambda^* = \lambda^{(t_{q^*})}_{q^*}$. Numerically, the BIC-based choice has tendency of overestimating $q$. For those overestimated, the corresponding choice of $\M{S}$ was very sparse in many rows, in that many rows have only one non-zero elements, for which the overall 0-norm values $||\V{\phi}^{(t)}_q||_0$ are not much different for different choices of $q$. Therefore, for choosing $q$, we modified the BIC criterion \eqref{eq:BIC} slightly to 
\begin{equation} \label{eq:mBIC}
mBIC(\V{\phi}^{(t)}_q, \lambda^{(t)}_q) = 2\mathcal{L}(\V{\phi}^{(t)}_q; \lambda^{(t)}_q) + q||\V{S}^{(t)}_q||_{2, 0}\log(N),
\end{equation}
where $\M{S}^{(t)}_q$ is the $\M{S}$ value of $\V{\phi}^{(t)}_q$, and $||\cdot||_{2, 0}$ is the $(2,0)$-matrix norm that counts the number of non-zero columns of a matrix.

\section{Simulated examples} \label{sec:sim}
This section present a numerical performance of the proposed variable selection approach with a number of simulated scenarios. We generate 27 simulated scenarios with different settings, each of which is characterized by an unique setting of simulation input parameters. For each scenario, we perform 25 simulation runs for replicated experiments. Each of the simulation runs starts with generating a dataset for a regression analysis with $p$ input variables, including $p_0$ inputs relevant to the response variable and $p-p_0$ irrelevant inputs. The data generation follows random sampling steps described below:
\begin{itemize}
	\item \textbf{Inputs:} noise variance $\sigma^2$, covariance parameter $\theta$, rank parameter $q$, $p$ and $p_0$.
	\item \textbf{Outputs:} $N$ records of input variables and response variable, $\M{X}, \V{y}$
	\item \textbf{Step 1.} Take an $N \times p$ input matrix $\M{X} = (\V{x}_1,\V{x}_2,\ldots, \V{x}_N)^T$ with each row $\V{x}_i \sim \Unif([0, 1]^{p})$ independently for $i = 1, \ldots, N$.
	\item \textbf{Step 2.} Sample the distance parameter $\M{S}$ as follows. Let $\M{A}$ denote a $p_0 \times q$ random matrix with each of the elements independently sampled from $\mathcal{N}(0, 1)$. Take the QR decomposition, $\M{A} = \M{O}_{p_0}\M{R}$, where $\M{O}_{p_0}$ is a $p_0 \times p_0$ orthonormal matrix and $\M{R}$ is a $p_0 \times q$ upper triangular matrix, and take a $q \times p_0$ submatrix $\M{O}_q$, made of the first $q$ rows of the orthonormal matrix for $q < p_0$. Sample a $q \times q$ diagonal matrix $\M{D}$ with with each diagonal element independently sampled from an inverse gamma distribution, $Gamma^{-1}(1,1)$. Set $\M{S}_q = \M{D} \M{O}_q$ and augment the $q \times p_0$ matrix $\M{S}_q$ to a $q \times p$ matrix by appending a $q \times (p-p_0)$ zero matrix. Randomly reorder the columns of the augmented matrix, which is set to $\M{S}$. 
	\item \textbf{Step 3.} Given $\M{S}$ from the previous step, we define a covariance function,
	\begin{equation} \label{eq:cov_num}
	c_S(\V{x}_i, \V{x}_j) = c_{exp}(d_S(\V{x}_i, \V{x}_j); \theta),
	\end{equation}
	where $c_{exp}$ is an exponential covariance function with variance parameter $\theta$. Sample $\V{y}|\V{X}, \M{S}, \theta, \sigma^2 \sim
	\mathcal{N}\left( \V{0}, \sigma^2\M{I} + \M{C}_{\M{S}, \theta} \right)$.
\end{itemize}
We fix $p=10$ and varied $p_0 \in \{3, 5, 7\}$. We also try different values of $\sigma^2 \in \{0.1^2, 0.3^2, 0.5^2\}$, while fixing the signal variance $\theta = 1$, which would create different signal-to-noise ratios. We also vary the rank parameter $q \in \{1, 2, 3\}$. The number of the possible combinations of the $p_0$, $\sigma^2$ and $q$ values is 27, and one unique setting serves as a simulation scenario. For each scenario, we perform 25 replicated simulation runs by generating 25 datasets, and the outcomes reported in this section are the statistics of the 25 outcomes, the mean and standard deviation. We first report an in-depth analysis of the outcomes from the proposed approach in Sections \ref{sec:sim_q} and \ref{sec:sim_proj}. Section \ref{sec:sim_hm} reports the comparison to three benchmark variable selection approaches, including the KL-divergence-based forward stepwise selection approach \citep[KL-F]{piironen2016projection}, KL-divergence-based sensitivity analysis \citep[KL-S]{paananen2019variable}, and variability-of-the-posterior-mean approach \citep[VAM]{paananen2019variable}. We have not included the comparison to the MCMC sampling approach \citep{savitsky2011variable}, mainly due to its computational slowness.

In the simulation study, we apply the $1$-norm sparsity penalty and set $T_{max}=100$, $\epsilon=10^{-3}$ and $\xi=10^{-6}$ for our approach. For all the three benchmark approaches, we use the BIC to choose the number of the variables selected.

\subsection{Analysis on the choice of tuning parameters} \label{sec:sim_q}
Our proposed approach has two tuning parameters, the rank parameter $q$ and the sparsity parameter $\lambda$. The rank $q$ determines the rank of $\M{Q}$ in the distance $d_S$ or equivalently the row size of the matrix $\M{S}$, and the sparsity parameter $\lambda$ determines the number of zero elements in the projection matrix $\M{S}$, which is related to the number of variables selected. We first analyze the choice of $q$ for the simulated scenarios in this section. We know the values of $q$ used to generate simulation scenarios, which are compared to the estimated $q^*$ achieved using the model selection approach described in Section \ref{sec:hyper}. The overall bias estimate of the estimation can be achieved by taking the mean of the observed $q^* - q$ values over $27 \times 25$ runs, which was -0.0148. If we drill down to the number, the percent with $q = q^*$ is 79.63\%, and the percent of $|q-q^*|\le 1$ is 95.56\%. Table \ref{tbl1} entails the percent splits. Since the accuracy did not depend significantly on $\sigma^2$ and $p_0$, we report the percents for each distinct $q$ and $q^*$ combination.
\begin{table}
	\centering
	\begin{tabular}{c c c c}
		\hline
		& $q^* = 1$ & $q^* = 2$ & $q^* =3$\\
		\hline
		$q = 1$ & 90.00\% & 10.00\% & 00.00\%  \\
		$q = 2$ & 14.44\% & 70.00\% & 15.56\%  \\  
		$q = 3$ & 13.33\% & 07.78\% & 78.89\%  \\ 
		\hline
	\end{tabular}
	\caption{Rank parameter $q$ versus estimated $q^*$ over 675 simulation runs} \label{tbl1}
\end{table}

\begin{figure}[ht!]
	\centering
	\includegraphics[width=1.1\textwidth]{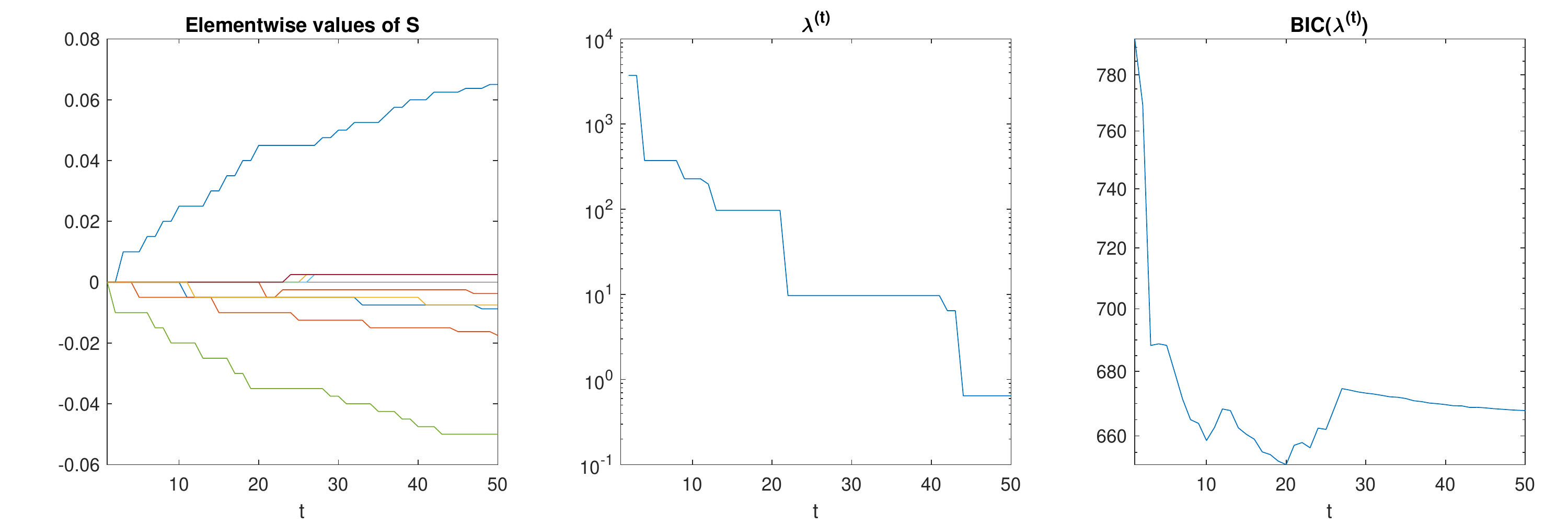}
	\caption{Solution path of the proposed \texttt{FSEG} approach for a simulation scenario with $p_0 = 5$, $q=1$ and $\sigma^2 = 0.5^2$. The left-most plot shows the values of the elements in the solution $\M{S}$ achieved by iteration $t$, the center plot shows the sparsity parameter values, $\lambda^{(t)}$, and the right plot shows the BIC value of the solution achieved at $t$. }
	\label{fig_sim}
\end{figure}

A solution path is also generated by the proposed \texttt{FSEG} algorithm. For a given choice $q^*$, the \texttt{FSEG} algorithm generates the solutions $\{\M{S}^{(t)}, \V{\phi}_C^{(t)}; t = 1,2,...\}$ of problem \eqref{opt:RML} for a decreasing sequence of $\lambda$ values, and we evaluate the BIC criterion \eqref{eq:BIC} of each of the solutions to choose the $\lambda$ value that minimizes the BIC criterion, which we denote by $\M{S}^{t_{q^*}}$ in Section \ref{sec:hyper}. Figure \ref{fig_sim} illustrates the solution path for a simulation scenario with $p_0 = 5$, $q=1$ and $\sigma^2 = 0.5^2$. The solution in the path that minimizes the BIC is achieved at $t=20$, and the number of non-zero columns in the solution $\M{S}^{(t_{q^*})}$ at $t=5$ is 5. This means the projection of the input variables, $\M{S}^{(t_{q^*})}\V{x}$, is a linear combination of the five variables corresponding to the five non-zero columns. The number of non-zero columns is equivalent to $p_0$, the number of variables used to generate the simulation data. We can also evaluate how the individual non-zero columns are compared to the ground truth, the variables used for the simulation data generation. The detailed report on this comparison will be discussed in Section \ref{sec:sim_hm}.

\subsection{Analysis of the estimated projection matrix} \label{sec:sim_proj}
\begin{figure}[ht!]
	\centering
	\includegraphics[width=1.0\textwidth]{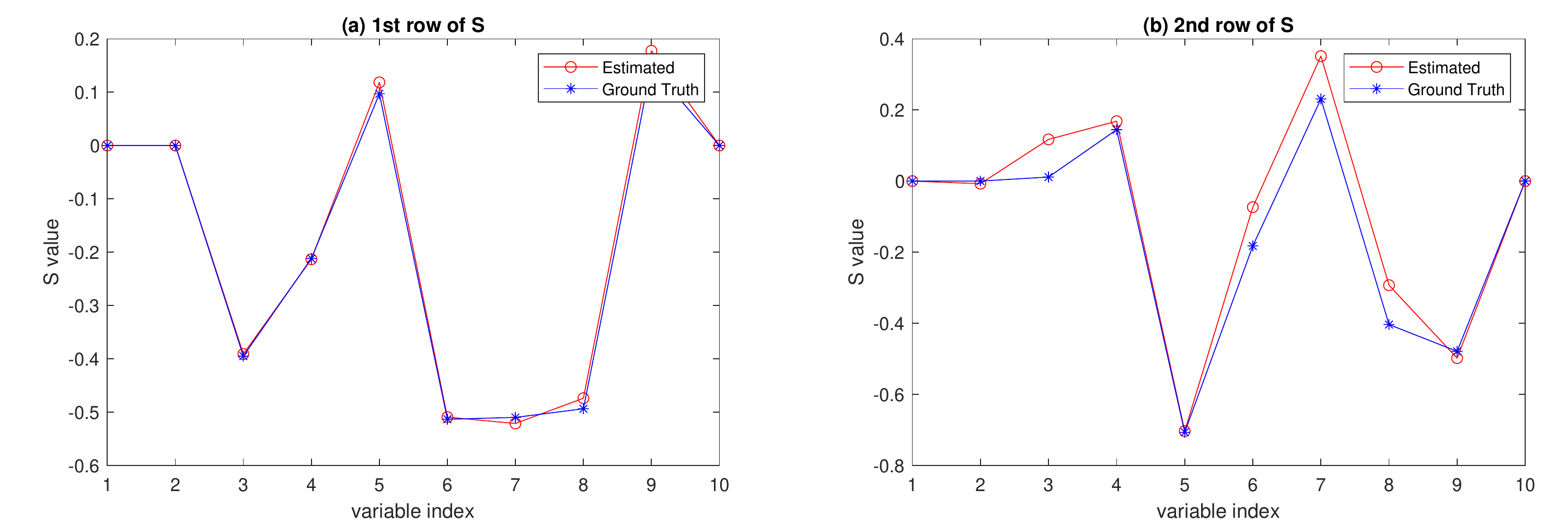}
	\caption{Comparison of the estimated $\M{S}^{(t_{q^*})}$ and the ground truth $\M{S}$ for $q=2$, $p_0 = 7$ and $\sigma^2 = 0.3^2$. }
	\label{fig_sim_S2}
\end{figure}

\begin{figure}[ht!]
	\centering
	\includegraphics[width=1.0\textwidth]{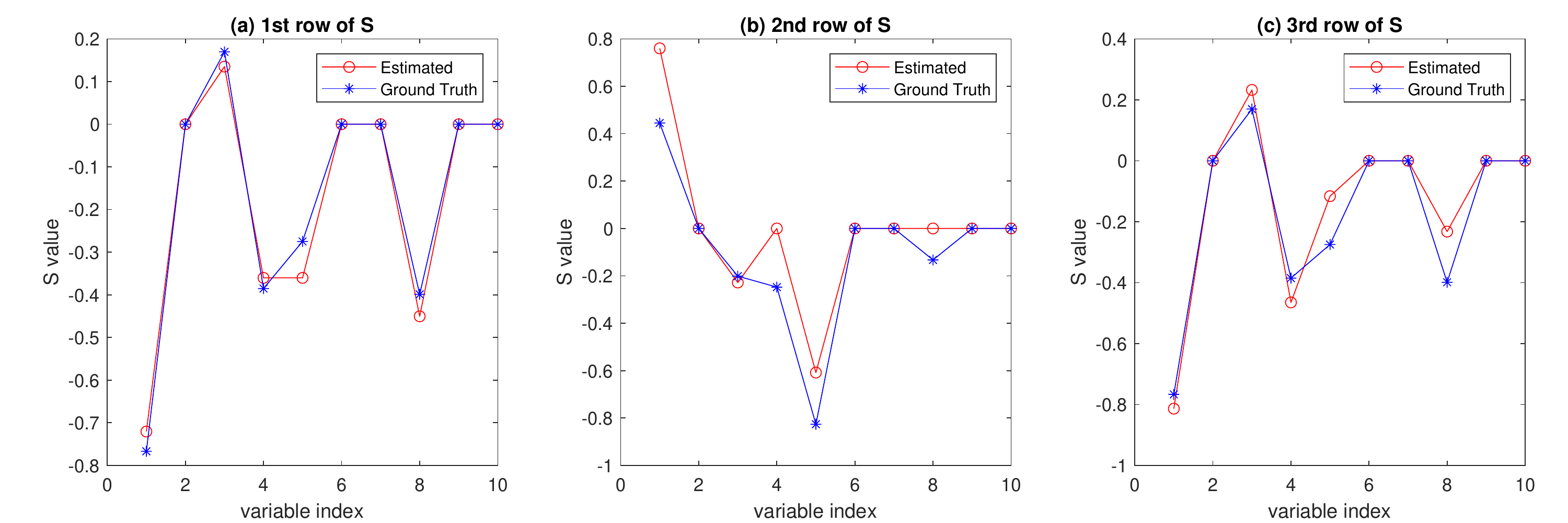}
	\caption{Comparison of the estimated $\M{S}^{(t_{q^*})}$ and the ground truth $\M{S}$ for $q=3$, $p_0 = 5$ and $\sigma^2 = 0.3^2$.}
	\label{fig_sim_S3}
\end{figure}

In this section, we analyze how the estimated projection matrix $\M{S}^{(t_{q^*})}$ is compared to the ground truth, i.e., the value of $\M{S}$ used for simulation data generation.  We calculated the Frobenius norm of the ground truth and the estimated one. Before the calculation, we reordered the rows of $\M{S}^{(t_{q^*})}$ so that the row-reordered matrix matches best to $\M{S}$. The row reordering is necessary for comparing the two matrices, because the row reorder does not make any change in both of the marginal likelihood and the sparsity penalty, so the $\M{S}^{(t_{q^*})}$ estimated by the proposed \texttt{FSEG} could have a different row ordering. Table \ref{tbl:S} summarizes the average and standard deviation of the Frobenius norm values over 25 simulation runs of each simulation scenario. Both of the mean and standard deviations did not vary much in $p_0$ and $\sigma^2$, but they changed significantly with $q$. For a higher rank $q$, there are more errors. This is because the size of $\M{S}$ is proportional to $q$, and there are many error sources involved for estimating a larger matrix. We also show $\M{S}^{(t_{q^*})}$ versus $\M{S}$ in Figures \ref{fig_sim_S2} and \ref{fig_sim_S3}.

\begin{table}
	\centering
	{\footnotesize
	\begin{tabular}{c | c c}
		\multicolumn{3}{c}{MSE of the estimated $\M{S}^{(t_{q^*})}$ for each simulation scenario} \\
		\hline
		($q$, $p_0$, $\sigma^2$) & average & standard deviation \\
		\hline
		(1,3,0.01) & 0.0000 & 0.0000 \\ 
		(1,3,0.09) & 0.0004 & 0.0011 \\ 
		(1,3,0.25) & 0.0003 & 0.0005 \\ 
		(1,5,0.01) & 0.0000 & 0.0000 \\ 
		(1,5,0.09) & 0.0000 & 0.0000 \\ 
		(1,5,0.25) & 0.0003 & 0.0004 \\ 
		(1,7,0.01) & 0.0000 & 0.0001 \\ 
		(1,7,0.09) & 0.0001 & 0.0004 \\ 
		(1,7,0.25) & 0.0020 & 0.0048 \\ 
		(2,3,0.01) & 0.0178 & 0.0172 \\ 
		(2,3,0.09) & 0.0094 & 0.0060 \\ 
		(2,3,0.25) & 0.0117 & 0.0091 \\ 
		(2,5,0.01) & 0.0149 & 0.0154 \\ 
		(2,5,0.09) & 0.0235 & 0.0161 \\ 
		(2,5,0.25) & 0.0127 & 0.0095 \\ 
		(2,7,0.01) & 0.0161 & 0.0123 \\ 
		(2,7,0.09) & 0.0201 & 0.0189 \\ 
		(2,7,0.25) & 0.0241 & 0.0186 \\ 
		(3,3,0.01) & 0.0224 & 0.0120 \\ 
		(3,3,0.09) & 0.0235 & 0.0177 \\ 
		(3,3,0.25) & 0.0190 & 0.0138 \\ 
		(3,5,0.01) & 0.0195 & 0.0074 \\ 
		(3,5,0.09) & 0.0237 & 0.0147 \\ 
		(3,5,0.25) & 0.0292 & 0.0140 \\ 
		(3,7,0.01) & 0.0252 & 0.0113 \\ 
		(3,7,0.09) & 0.0315 & 0.0140 \\ 
		(3,7,0.25) & 0.0220 & 0.0141 \\ 
		\hline
		Overall & 0.0137 & 0.0092 \\ 
		\hline
	\end{tabular}
	\caption{Averages and standard deviations of the MSE of the estimated $\M{S}^{(t_{q^*})}$ over 25 runs of each simulation scenario. The first column of the table shows the simulation input parameter values used to generate each simulation scenario. }
}
	\label{tbl:S}
\end{table}

\subsection{Hit-and-miss of relevant variables} \label{sec:sim_hm}
For each simulation scenario, we also analyze the variables identified by the proposed approach, which are compared to the set of $p_0$ relevant variables used in the data generation procedure (regarded as the ground truth). The variables identified by the proposed approach are achieved as the variables corresponding to the non-zero columns in the estimated $\M{S}^{(t_{q^*})}$. Let $\M{A}$ denote the set of $p_0$ relevant variables used in the simulation data generation, and let $\M{\hat{A}}$ denote the set of the variables identified by the proposed approach. We count the false positive rate (FNR) and the faulty positive rate (FPR) error of $\hat{A}$ versus $A$. 
\begin{equation*}
\begin{split}
& \mbox{FNR} = \frac{|A - \hat{A}|}{|A|} \\
& \mbox{FPR} = \frac{|\hat{A} - A|}{10-|A|},
\end{split}
\end{equation*}
where $|\cdot|$ denotes the set cardinality, and $-$ is a set difference operator. The FNR and FPR values are calculated, and the means and standard deviations of the two values are taken over 25 simulation runs for each of the simulated scenarios. The same evaluations are performed for some chosen benchmark methods, including the KL-divergence-based forward stepwise selection approach \citep[KL-F]{piironen2016projection}, KL-divergence-based sensitivity analysis \citep[KL-S]{paananen2019variable}, and variability-of-the-posterior-mean approach \citep[VAM]{paananen2019variable}. The individual statistics are reported in Table \ref{tbl:MD} and Table \ref{tbl:FD} for comparison. We summarize the outcomes in a graphical plot showing the average FPR values versus the corresponding average true positive rates (TPR = 1 - FNR), borrowing the ROC plotting style popularly used to present machine learning algorithms. Typically, if the FPR value was lowered, the TPR value would decrease. The graphical plot would show what approaches provide better trade-offs in between the FPR and TPR values. The overall FPR values are pretty comparable among all the compared methods, which have shown more differences in the TPR values. The proposed approach achieves the highest TPR values (i.e. lowest FNR values) among the compared methods for most of the compared scenarios. In particular, the proposed approach exhibits a larger gap to the benchmark approaches for the scenarios with high noise variance $\sigma^2$. The proposed approach is pretty robust to high noises. 

\begin{figure}[ht!]
	\centering
	\includegraphics[width=0.8\textwidth]{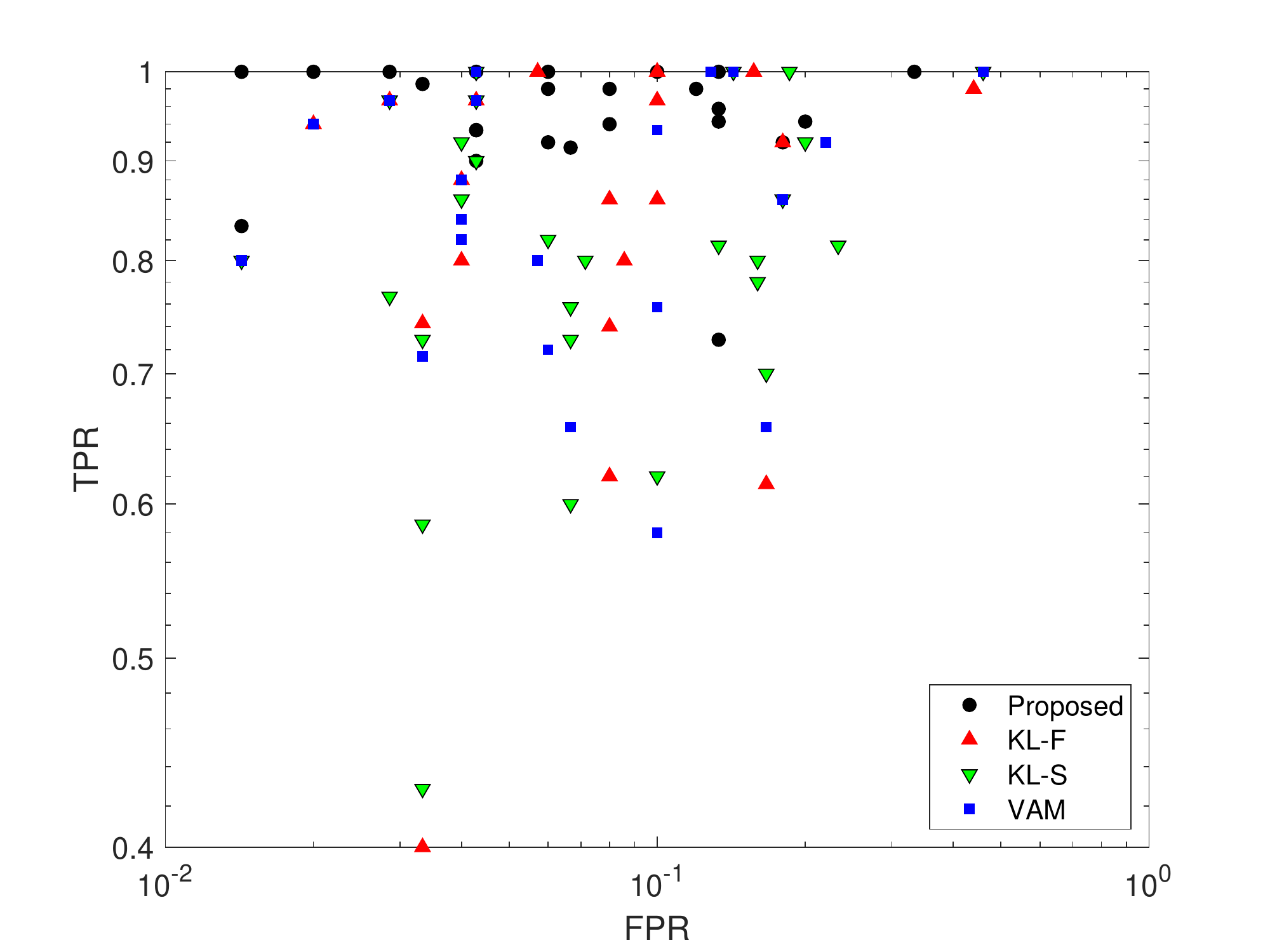}
	\caption{Receiver operating characteristic (ROC) curves of the proposed approach compared with those from benchmark approaches.  }
	\label{fig_sim_roc}
\end{figure}

\begin{table}
	\centering
	{\footnotesize
	\begin{tabular}{c | c c c c}
		\multicolumn{5}{c}{Average and Standard Deviation of FNRs by Scenario} \\
		\hline
		($q$, $p_0$, $\sigma^2$) & Proposed & KL-F & KL-S & VAM \\
		\hline
		(1,3,0.01) & 0.00 (0.00) & 0.03 (0.11) & 0.10 (0.22) & 0.07 (0.21) \\ 
		(1,3,0.09) & 0.07 (0.21) & 0.13 (0.17) & 0.23 (0.22) & 0.17 (0.18) \\ 
		(1,3,0.25) & 0.07 (0.14) & 0.20 (0.17) & 0.20 (0.17) & 0.20 (0.17) \\ 
		(1,5,0.01) & 0.00 (0.00) & 0.12 (0.17) & 0.22 (0.29) & 0.16 (0.26) \\ 
		(1,5,0.09) & 0.02 (0.06) & 0.26 (0.27) & 0.20 (0.25) & 0.28 (0.27) \\ 
		(1,5,0.25) & 0.02 (0.06) & 0.38 (0.30) & 0.38 (0.36) & 0.42 (0.35) \\ 
		(1,7,0.01) & 0.03 (0.06) & 0.26 (0.21) & 0.19 (0.15) & 0.26 (0.21) \\ 
		(1,7,0.09) & 0.06 (0.10) & 0.34 (0.19) & 0.41 (0.26) & 0.34 (0.22) \\ 
		(1,7,0.25) & 0.09 (0.15) & 0.60 (0.31) & 0.57 (0.33) & 0.59 (0.33) \\ 
		(2,3,0.01) & 0.00 (0.00) & 0.00 (0.00) & 0.00 (0.00) & 0.00 (0.00) \\ 
		(2,3,0.09) & 0.00 (0.00) & 0.03 (0.11) & 0.03 (0.11) & 0.03 (0.11) \\ 
		(2,3,0.25) & 0.10 (0.32) & 0.03 (0.11) & 0.03 (0.11) & 0.03 (0.11) \\ 
		(2,5,0.01) & 0.06 (0.19) & 0.14 (0.31) & 0.14 (0.31) & 0.14 (0.31) \\ 
		(2,5,0.09) & 0.00 (0.00) & 0.06 (0.10) & 0.08 (0.14) & 0.06 (0.10) \\ 
		(2,5,0.25) & 0.02 (0.06) & 0.08 (0.10) & 0.08 (0.10) & 0.08 (0.10) \\ 
		(2,7,0.01) & 0.01 (0.05) & 0.33 (0.23) & 0.24 (0.26) & 0.29 (0.24) \\ 
		(2,7,0.09) & 0.00 (0.00) & 0.31 (0.28) & 0.27 (0.23) & 0.29 (0.28) \\ 
		(2,7,0.25) & 0.06 (0.10) & 0.33 (0.27) & 0.27 (0.26) & 0.29 (0.27) \\ 
		(3,3,0.01) & 0.00 (0.00) & 0.00 (0.00) & 0.00 (0.00) & 0.00 (0.00) \\ 
		(3,3,0.09) & 0.00 (0.00) & 0.00 (0.00) & 0.00 (0.00) & 0.00 (0.00) \\ 
		(3,3,0.25) & 0.17 (0.36) & 0.20 (0.36) & 0.20 (0.36) & 0.20 (0.36) \\ 
		(3,5,0.01) & 0.08 (0.25) & 0.02 (0.06) & 0.00 (0.00) & 0.00 (0.00) \\ 
		(3,5,0.09) & 0.00 (0.00) & 0.14 (0.25) & 0.14 (0.25) & 0.12 (0.19) \\ 
		(3,5,0.25) & 0.08 (0.25) & 0.20 (0.34) & 0.18 (0.32) & 0.18 (0.32) \\ 
		(3,7,0.01) & 0.00 (0.00) & 0.23 (0.24) & 0.19 (0.21) & 0.24 (0.25) \\ 
		(3,7,0.09) & 0.27 (0.44) & 0.39 (0.35) & 0.30 (0.37) & 0.34 (0.37) \\ 
		(3,7,0.25) & 0.04 (0.10) & 0.40 (0.32) & 0.40 (0.32) & 0.43 (0.34) \\ 
		\hline
		Overall & 0.05 (0.108) & 0.19 (0.198) & 0.19 (0.208) & 0.19 (0.205) \\ 
		\hline
	\end{tabular}
	\caption{Averages and standard deviations of the FNR values over 25 runs of each simulation scenario. The first column of the table shows the simulation input parameter values used to generate each simulation scenario. For the second to the fifth columns, the non-bracketed numbers are the averages, and the non-bracketed ones are the standard deviations.}}
	\label{tbl:MD}
\end{table}

\begin{table}
	{\footnotesize
	\centering
	\begin{tabular}{c | c c c c}
		\multicolumn{5}{c}{Average and Standard Deviation of FPRs by Scenario} \\
		\hline
		($q$, $p_0$, $\sigma^2$) & Proposed & KL-F & KL-S & VAM \\
		\hline
		(1,3,0.01) & 0.00 (0.00) & 0.10 (0.15) & 0.04 (0.07) & 0.10 (0.18) \\ 
		(1,3,0.09) & 0.04 (0.10) & 0.00 (0.00) & 0.03 (0.06) & 0.00 (0.00) \\ 
		(1,3,0.25) & 0.04 (0.07) & 0.00 (0.00) & 0.01 (0.05) & 0.01 (0.05) \\ 
		(1,5,0.01) & 0.02 (0.06) & 0.04 (0.08) & 0.16 (0.31) & 0.04 (0.08) \\ 
		(1,5,0.09) & 0.12 (0.27) & 0.08 (0.19) & 0.16 (0.23) & 0.06 (0.13) \\ 
		(1,5,0.25) & 0.08 (0.10) & 0.08 (0.19) & 0.10 (0.19) & 0.10 (0.19) \\ 
		(1,7,0.01) & 0.00 (0.00) & 0.03 (0.11) & 0.23 (0.35) & 0.00 (0.00) \\ 
		(1,7,0.09) & 0.13 (0.23) & 0.00 (0.00) & 0.03 (0.11) & 0.07 (0.21) \\ 
		(1,7,0.25) & 0.07 (0.14) & 0.03 (0.11) & 0.03 (0.11) & 0.00 (0.00) \\ 
		(2,3,0.01) & 0.03 (0.09) & 0.10 (0.19) & 0.04 (0.14) & 0.04 (0.14) \\ 
		(2,3,0.09) & 0.04 (0.07) & 0.03 (0.09) & 0.03 (0.09) & 0.03 (0.09) \\ 
		(2,3,0.25) & 0.04 (0.07) & 0.04 (0.10) & 0.04 (0.10) & 0.04 (0.10) \\ 
		(2,5,0.01) & 0.08 (0.19) & 0.10 (0.19) & 0.18 (0.35) & 0.18 (0.35) \\ 
		(2,5,0.09) & 0.06 (0.10) & 0.02 (0.06) & 0.04 (0.08) & 0.02 (0.06) \\ 
		(2,5,0.25) & 0.06 (0.10) & 0.18 (0.30) & 0.20 (0.30) & 0.22 (0.36) \\ 
		(2,7,0.01) & 0.03 (0.11) & 0.00 (0.00) & 0.07 (0.14) & 0.03 (0.11) \\ 
		(2,7,0.09) & 0.33 (0.27) & 0.00 (0.00) & 0.03 (0.11) & 0.03 (0.11) \\ 
		(2,7,0.25) & 0.20 (0.32) & 0.00 (0.00) & 0.07 (0.21) & 0.00 (0.00) \\ 
		(3,3,0.01) & 0.00 (0.00) & 0.16 (0.25) & 0.19 (0.31) & 0.13 (0.25) \\ 
		(3,3,0.09) & 0.01 (0.05) & 0.06 (0.18) & 0.14 (0.30) & 0.14 (0.30) \\ 
		(3,3,0.25) & 0.01 (0.05) & 0.09 (0.18) & 0.07 (0.14) & 0.06 (0.14) \\ 
		(3,5,0.01) & 0.18 (0.30) & 0.44 (0.40) & 0.46 (0.34) & 0.46 (0.38) \\ 
		(3,5,0.09) & 0.10 (0.19) & 0.08 (0.19) & 0.04 (0.08) & 0.04 (0.08) \\ 
		(3,5,0.25) & 0.06 (0.13) & 0.04 (0.08) & 0.06 (0.10) & 0.04 (0.08) \\ 
		(3,7,0.01) & 0.13 (0.17) & 0.00 (0.00) & 0.13 (0.23) & 0.10 (0.22) \\ 
		(3,7,0.09) & 0.13 (0.23) & 0.17 (0.36) & 0.17 (0.24) & 0.17 (0.28) \\ 
		(3,7,0.25) & 0.13 (0.23) & 0.00 (0.00) & 0.07 (0.14) & 0.00 (0.00) \\ 
		\hline 
		Overall & 0.08 (0.135) & 0.07 (0.127) & 0.10 (0.180) & 0.08 (0.144) \\ 
		\hline
	\end{tabular}
	\caption{Averages and standard deviations of the FPR values over 25 runs of each simulation scenario. The first column of the table shows the simulation input parameter values used to generate each simulation scenario. For the second to the fifth columns, the non-bracketed numbers are the averages, and the non-bracketed ones are the standard deviations.}}
	\label{tbl:FD}
\end{table}

\section{Real example: environmental corrosion analysis} \label{sec:data}
This section presents the application of the proposed variable selection approach to identify the environmental factors most influential to metal corrosion. The outcome will be exploited to design an accelerated corrosion testing protocol using a custom environmental chamber that can simulate real-world conditions including temperature, relative humidity, salt water
spray, background gases, and artificial sunlight. Developing the protocol would require two preliminary steps: first identifying the control factors and then calibrating the control factor
levels. The benefit of this exercise will be to reduce the number of factors to account for when conducting an experiment in the laboratory test chamber that produces similar metal
corrosion to that occurring in a natural environment. Pre-selecting a subset of more influential factors is highly desirable for a more efficient design of the accelerated corrosion test protocol.

\begin{figure}[ht!]
	\centering
	\includegraphics[width=0.8\textwidth]{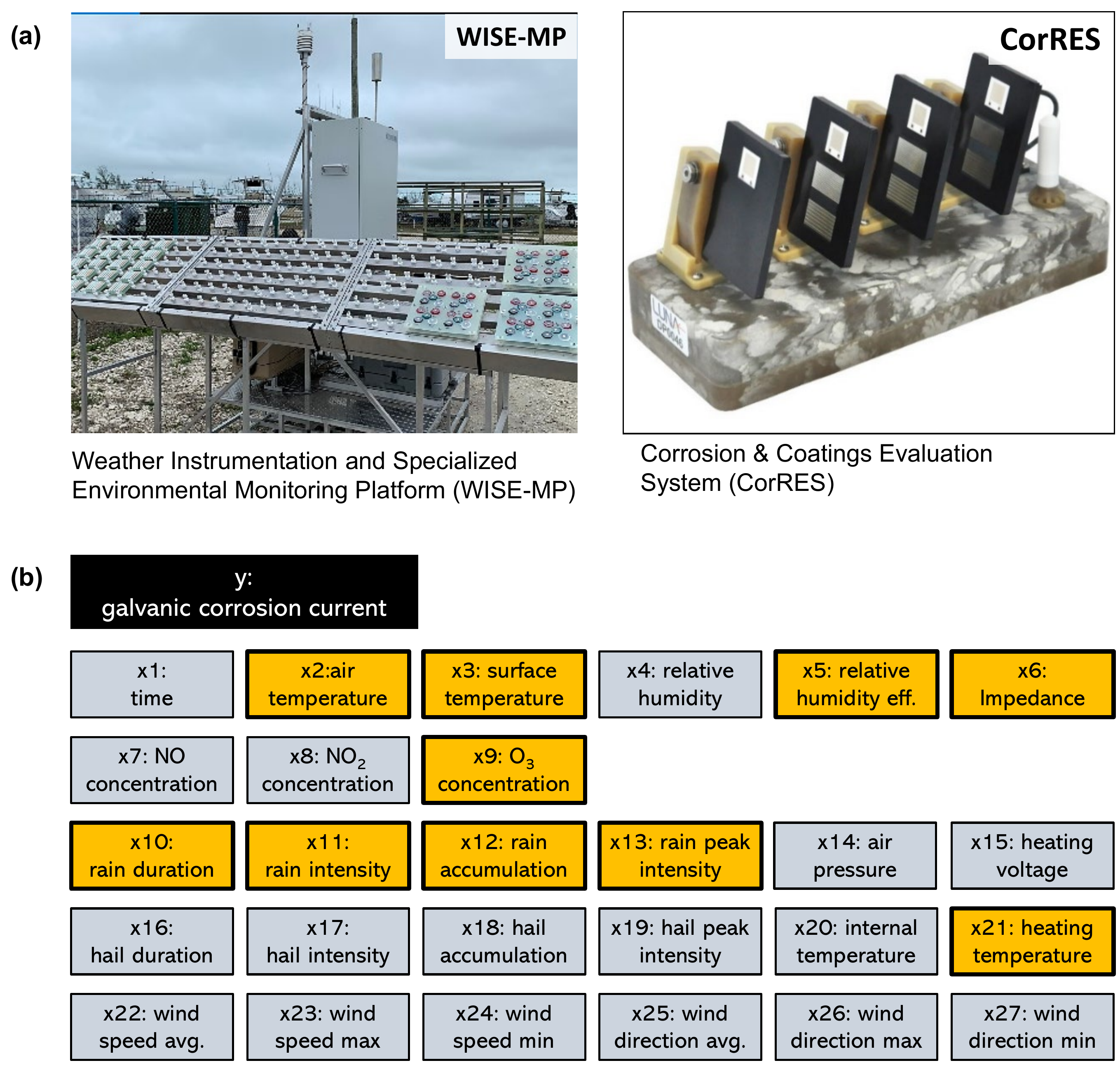}
	\caption{Data collection systems for a metal corrosion study. (a) shows a picture of the WISE-MP system, and (b) shows the measurements taken from the WISE-MP and CorRES.}
	\label{fig_data}
\end{figure}

For the variable selection, the U.S. Air Force deployed two measurement systems to collect necessary data, the Corrosion \& Coatings Evaluation System ({CorRES}\textsuperscript{\texttrademark}), and the Weather Instrumentation and Specialized Environmental Monitoring Platform ({WISE-MP}), shown in Figure \ref{fig_data}-(a). Both systems were placed at a test site operated by the U.S. Naval Research Laboratory in Key West FL. The two measurement systems produced the periodic measurements of 27 environmental factors that potentially affect atmospheric corrosion of aluminum alloy (AA) specimens attached on the sensing systems, including different temperature measurements, relative humidity, concentrations of several corrosive gases, and other weather conditions such as the intensities and durations of rain, hail and wind. A complete list of the factors can be found in Figure \ref{fig_data}-(b). The galvanic corrosion current flowing through the AA specimen was also measured to quantify the degree of corrosion of the specimen. In total, 18,016 records of the environmental factors and corrosion current measurements were collected over a 3 month period from May 31 2019 to August 22 2019.  

\begin{figure}[ht!]
	\centering
	\includegraphics[width=\textwidth]{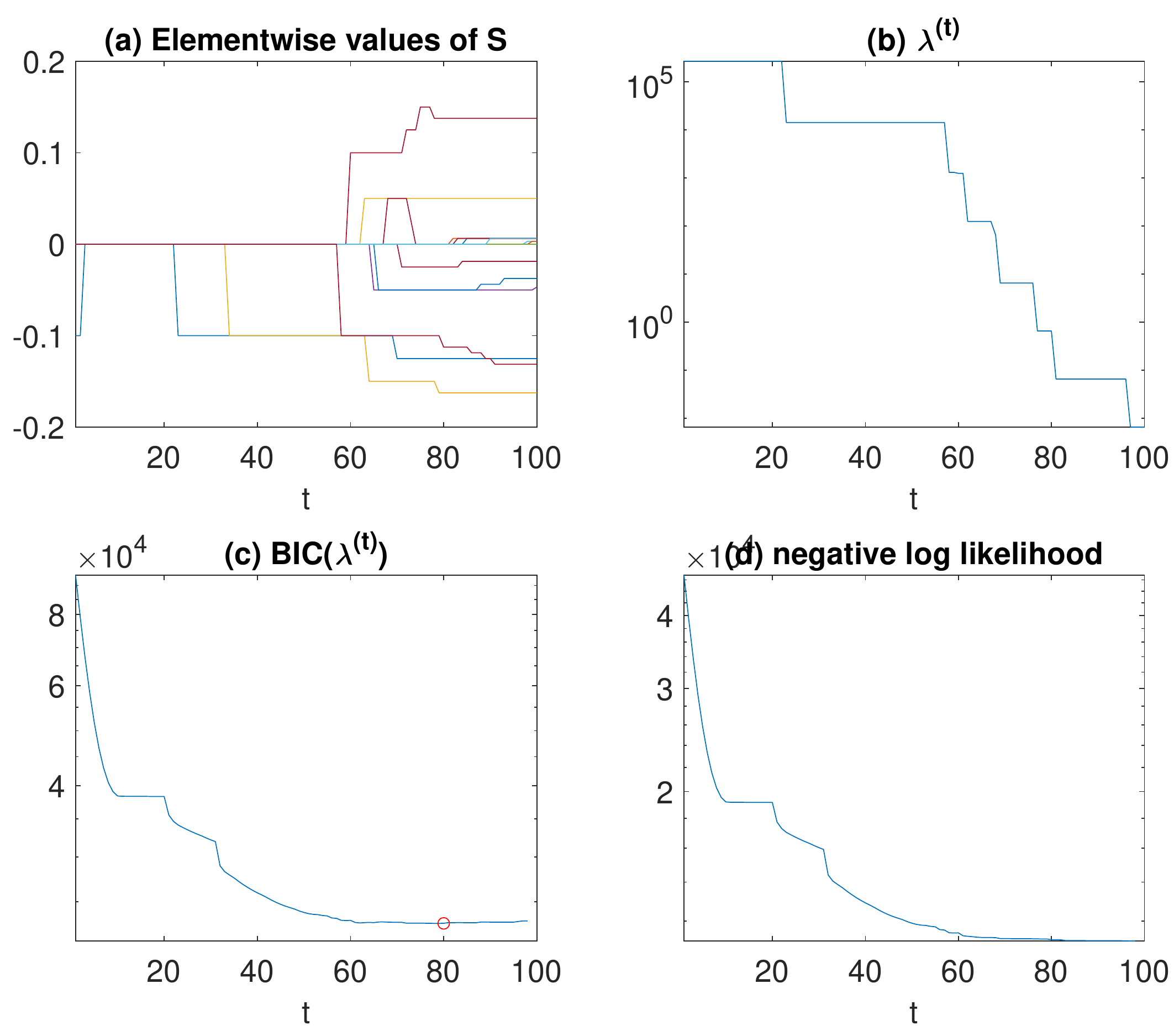}
	\caption{Solution path of the proposed \texttt{FSEG} approach for the WISE-MP corrosion dataset. The $q=2$ is chosen using the model selection procedure in Section \ref{sec:hyper}. (a) shows the solution path generated by the \texttt{FSEG}, (b) shows the corresponding trace of the sparsity parameter values applied, (c) shows the BIC versus iteration $t$ with the lowest BIC value circled,  and (d) shows the negative log likelihood value versus iteration $t$.}
	\label{fig_real}
\end{figure}

The corrosion currents were related to the environmental factors through a GP regression model, and our approach was applied to select a subset of the 27 environmental factors that contribute most to accurate GP modeling. To evaluate the outcome of the GP modeling and variable selection, we randomly split the 18,016 records into two sets, a training set for training the GP regression with the proposed variable selection and a test set for evaluating the outcome. The split ratio was eight to one, eight for the training set and one for the testing set. The training set was composed of 14,411 records, for which the marginal likelihood calculation would take a very long time. We used an approximation to the marginal likelihood and the corresponding GP regression, based on the patchwork Kriging \citep{park2018patchwork}. In the approximation scheme, the data is partitioned into $K$ subsets, $\{(\V{X}_k, \V{y}_k); k = 1,\ldots, K\}$, and the approximate likelihood is defined as a sum of the likelihoods over the subsets,
\begin{equation*}
2\mathcal{L}_a(\M{S}, \V{\phi}_C) = \sum_{k=1}^K \left[\V{y}_k^T (\sigma^2\M{I} + \M{C}^{(k)}_{\M{S}, \V{\theta}})^{-1} \V{y}_k + \log |\sigma^2\M{I} + \M{C}^{(k)}_{\M{S}, \V{\theta}}|\right],
\end{equation*}
where $\M{C}^{(k)}_{\M{S}, \V{\theta}}$ is the covariance function evaluated for the $k$th subset, $\V{X}_k$. We used $K=40$, and the covariance function used in the simulation study is applied. We set $\epsilon = 10^{-2}$, $\xi = 10^{-6}$ and $T_{max} = 1,000$ for the proposed \texttt{FSEG}, and the $q$ and $\lambda$ were chosen by the model selection described in Section \ref{sec:hyper}. The chosen value of $q$ is 2.  Figure \ref{fig_real} shows the solution path for $q=2$ and the corresponding lambda values over the first 100 iterations. The lowest BIC value was achieved at iteration $t=80$, for which the sparsity parameter $\lambda$ was 0.6516. The solution achieving the lowest BIC value was selected as the final estimate of the GP parameters, $\V{\phi}_C$ and $\V{S}$. The estimate of $\M{S}$ provided the relevance of 27 variables to the galvanic corrosion. According to the estimate, ten among 27 variables are relevant to the corrosion rate. The ten relevant variables are highlighted with yellow colors in Figure \ref{fig_data}-(b), including air temperature, surface temperature, heating temperature, effective relative humidity, electrochemical impedance, concentration of $O_3$ and four rain related weather conditions. 

We evaluated the outcome of the variable selection and the corresponding GP model quantitatively and qualitatively. For the quantitative judgment, we fit two regression models to the training dataset, one GP regression model with a full set of the 27 environmental factors and another GP model with the ten selected factors, and we compared the prediction accuracies of the two models in terms of their posterior mean and variance estimates. For comparison of prediction accuracy, we calculated two performance metrics on the test data, denoted by $\{(x_t, y_t): t=1,\dots,T\}$, where $T$ is the test set size. Let $\mu_t$ and $\sigma^2_t$ denote the estimated posterior mean and variance at location $x_t$. The first measure is the mean squared error (MSE)
\begin{equation}
\textrm{MSE} = \frac{1}{T} \sum_{t=1}^T (y_t - \mu_t)^2,
\end{equation}
which measures the accuracy of the mean prediction $\mu_t$ at location $x_t$. The second measure is the negative log predictive density (NLPD)
\begin{equation}
\textrm{NLPD} = \frac{1}{T} \sum_{t=1}^T\left[ \frac{(y_t - \mu_t)^2}{2\sigma_t^2} + \frac{1}{2} \log (2\pi \sigma_t^2) \right].
\end{equation}
The NLPD quantifies the degree of fitness of the estimated predictive distribution $\mathcal{N}(\mu_t, \sigma_t^2)$ for the test data. These two criteria are used broadly in the GP regression literature. A smaller value of MSE or NLPD indicates better performance. Table \ref{tbl:real} compares the MSE and NLPD values. The reduced model with the ten selected inputs performed better in both the MSE and the NLPD. This means that the mean and posterior variance estimates with the reduced model better fit to the test data, so the ten selected variables correlate well to the corrosion current. 

\begin{table}
	\centering
	\begin{tabular}{l c c}
		\hline
		& MSE    & NLPD\\
		\hline
		full GP with all 27 factors 			 & 0.0475 & -0.6035   \\
		reduced GP with the ten selected factors & 0.0434 & -1.0865 \\  
		\hline
	\end{tabular}
	\caption{Comparison of the full GP model and the reduced GP model for the environmental corrosion data} \label{tbl:real}
\end{table}

We also evaluated the ten selected variables qualitatively based on a corrosion scientist's expert's judgment. The ten selected variables are regarded as important factors influencing environmental corrosion. Temperature and relative humidity have been identified as major drivers of corrosion in many existing works \citep{friedersdorf2019electrochemical, zheng2009atmospheric}. The ozone $O_3$ level and   electrochemical impedance measured using an AC signal at a high frequency (25 KHz) are among the factors popularly studied \citep{sae2019}. Ozone is a strong oxidizer that can lead to significant corrosion and material degradation at a high exposure level, and the impedance can be correlated to the amount of pollutants on the surface of a metal specimen \citep{friedersdorf2019electrochemical}. The effect of rain on the corrosion behavior of aluminum is more complicated because it can both reduce corrosion by washing inorganic pollutants off the surface as well as increase corrosion by scrubbing gases out of the air, becoming acid rain.  It is generally accepted that rain reduces the corrosion rate on aluminum \citep[page 245]{vargel2020corrosion}. The evidence for this is that outdoor samples covered from the rain have a higher corrosion rate than ones left out in the rain. 

\section{Conclusion} \label{sec:conc}
We presented a novel variable selection approach for GP regression, based on a sparse projection of input variables. The approach can be thought of as a generalization of the automatic relevance determination with a sparsity prior. The major distinctions from the existing approaches are that our approach estimates the sparse projection matrix jointly with other covariance parameters through a marginal likelihood maximization with a sparsity regularization on the projection matrix, while many existing approaches use slow MCMC samplings. In our initial numerical trials, we have tried a simple gradient descent and a quasi Newton Raphson algorithm, but they did not give satisfactory outcomes. In particular, the projection matrix tends to be very dense even with a large sparsity penalty. We proposed a forward stagewise regression with embedded gradient descent steps. The numerical approach is an extension of the existing forward stagewise Lasso for a non-convex objective function. We provided some convergence properties. The proposed approach worked successfully for many simulated scenarios, and its variable selection accuracy outperformed some benchmark approaches for most of the simulated scenarios. The approach was also applied to an important problem of identifying environmental factors that affect an atmospheric corrosion of a metal alloy, and its variable selection outcome is evaluated quantitatively and qualitatively.  

\if0\blind
{
\section*{Acknowledgment} 
We acknowledge support for this work from the prime contract of the U.S. Federal Government, Contract No. FA8650-15-D-5405. 
} \fi

\bibliographystyle{agsm}
\bibliography{GPDR}
\end{document}